\documentclass[10pt,twocolumn,letterpaper]{article}

\usepackage{cvpr}
\usepackage{times}
\usepackage{epsfig}
\usepackage{graphicx}
\usepackage{amsmath}
\usepackage{amssymb}
\usepackage[normalem]{ulem}
\usepackage{hhline}
\usepackage{caption}
\usepackage{subcaption}
\usepackage[accsupp]{axessibility}
\useunder{\uline}{\ul}{}

\newcommand{\multilinecomment}[1]{}

\usepackage[pagebackref=true,breaklinks=true,letterpaper=true,colorlinks,bookmarks=false]{hyperref}

\begin{document}

\title{EFaR 2023: Efficient Face Recognition Competition}

\author{
\begin{minipage}{0.95\textwidth}
\small
\renewcommand{\baselinestretch}{1.15}
\centering
Jan Niklas Kolf$^{1,2,*}$, Fadi Boutros$^{1,*}$, Jurek Elliesen$^{1,2,*}$, Markus Theuerkauf$^{1,2,*}$, Naser Damer$^{1,2,*}$\\
Mohamad Alansari$^{3,+}$, Oussama Abdul Hay$^{3,+}$, Sara Alansari$^{3,+}$, Sajid Javed$^{3,+}$, Naoufel Werghi$^{3,+}$, Klemen Grm$^{4,+}$,\\
Vitomir Štruc$^{4,+}$, Fernando Alonso-Fernandez$^{5,+}$, Kevin Hernandez Diaz$^{5,+}$, Josef Bigun$^{5,+}$, Anjith George$^{6,+}$,\\
Christophe Ecabert$^{6,+}$, Hatef Otroshi Shahreza$^{6,7,+}$, Ketan Kotwal$^{6,+}$, Sébastien Marcel$^{6,8,+}$, Iurii Medvedev$^{9,+}$,\\
Bo Jin$^{9,+}$, Diogo Nunes$^{9,+}$, Ahmad Hassanpour$^{10,+}$, Pankaj Khatiwada$^{10,+}$, Aafan Ahmad Toor$^{10,+}$, Bian Yang$^{10,+}$\\
$^{1}$Fraunhofer Institute for Computer Graphics Research IGD, Germany 
$^{2}$TU Darmstadt, Germany\\
$^{3}$Department of Electrical and Computer Engineering, Khalifa University, Abu Dhabi, United Arab Emirates\\
$^{4}$ Laboratory for Machine Intelligence, Faculty of Electrical Engineering, University of Ljubljana, Slovenia $^{5}$ Halmstad University, Sweden $^{6}$ Idiap Research Institute, Martigny, Switzerland $^{7}$ École Polytechnique Fédérale de Lausanne (EPFL), Lausanne, Switzerland. $^{8}$ Université de Lausanne (UNIL), Lausanne, Switzerland. $^{9}$ Institute of Systems and Robotics, University of Coimbra, Coimbra, Portugal $^{10}$ eHealth and Welfare Security Group, Department of Information Security and Communication Technology, Norwegian University of Science and Technology, Norway \\
$^{*}$Competition organizer $^{+}$Competition participant\\
Email: {jan.niklas.kolf@igd.fraunhofer.de}
\end{minipage}}

\maketitle
\thispagestyle{empty}

\begin{abstract}
   This paper presents the summary of the Efficient Face Recognition Competition (EFaR) held at the 2023 International Joint Conference on Biometrics (IJCB 2023). The competition received $17$ submissions from $6$ different teams. To drive further development of efficient face recognition models, the submitted solutions are ranked based on a weighted score of the achieved verification accuracies on a diverse set of benchmarks, as well as the deployability given by the number of floating-point operations and model size. The evaluation of submissions is extended to bias, cross-quality, and large-scale recognition benchmarks. Overall, the paper gives an overview of the achieved performance values of the submitted solutions as well as a diverse set of baselines. The submitted solutions use small, efficient network architectures to reduce the computational cost, some solutions apply model quantization. An outlook on possible techniques that are underrepresented in current solutions is given as well.
\end{abstract}
\vspace{-6mm}


\section{Introduction}
\vspace{-2mm}
Biometrics uses the behavioral and physical characteristics of a person for recognition \cite{DBLP:journals/tcsv/JainRP04}. These systems are increasingly used in everyday life. Modern smartphones are equipped with front cameras that enable the use of the face as a biometric modality \cite{DBLP:journals/psychnology/SpolaorLMCGS16}. High-performing face recognition (FR) systems were specifically enabled with the breakthrough of deep neural networks (DNN) and deep convolutional neural networks (CNN) for computer vision tasks \cite{DBLP:conf/cvpr/BoutrosDKK22, arcface, DBLP:conf/nips/KrizhevskySH12, DBLP:journals/ijon/WangD21a}. Consequently, the performance of state-of-the-art (SOTA) FR models has been greatly improved, even under challenging conditions \cite{DBLP:journals/corr/abs-2103-01716, arcface, 10007963}. Likewise, SOTA FR models surpass human performance on various benchmarks \cite{DBLP:journals/ijon/WangD21a}. 

However, the increased performance of DNNs is associated with high computational complexity, which makes it difficult to apply the FR models in resource-restricted domains such as embedded devices or smartphones \cite{DBLP:conf/icpr/BoutrosDK22, DBLP:conf/iccvw/DengGZDLS19, DBLP:journals/air/Martinez-DiazNM21}. To utilize DNN on embedded devices, the computational complexity, measured as the number of floating-point operations (FLOPs), and the memory footprint (in MB) need to meet certain criteria. An indication of the targeted memory footprint of a DNN that can be deployed on embedded devices is the ICCV 2019 Workshop challenge on Eye Tracking organized by Facebook \cite{DBLP:conf/iccvw/2019}. There, the targeted model footprint is set to 1 MB. A targeted upper bound of FLOPs for lightweight FR models is given by the ICCV 2019 Lightweight Face Recognition Challenge \cite{DBLP:conf/iccvw/DengGZDLS19}. Its lightweight FR model track defined models with at most 1G FLOPs as lightweight models. To develop FR models that meet these criteria and address the challenges, different approaches are used.

One approach is to design lightweight DNN models \cite{DBLP:conf/icb/BoutrosDFKK21}, where a small and efficient network architecture is developed that reduces the computational effort in comparison to larger and more complex DNN \cite{DBLP:journals/access/BoutrosSKDKK22}. Recent advancements in lightweight DNN are MobileNet \cite{DBLP:conf/cvpr/SandlerHZZC18}, SqueezeNet \cite{DBLP:journals/corr/IandolaMAHDK16_squeezenet}, VarGNet \cite{DBLP:journals/corr/abs-1907-05653_vargnet}, MixNet
\cite{DBLP:conf/bmvc/TanL19_mixnet}, ShuffleNet, \cite{DBLP:conf/eccv/MaZZS18} and GhostNet \cite{DBLP:conf/cvpr/HanW0GXX20_GhostNet}. For FR, these architectures developed for common computer vision tasks are increasingly used and adapted \cite{DBLP:conf/ccbr/ChenLGH18, DBLP:conf/iccvw/Martinez-DiazLV19}.  One of the earliest works to use lightweight DNN for FR was MobileFaceNet \cite{DBLP:conf/ccbr/ChenLGH18} which utilized MobileNetV2 \cite{DBLP:conf/cvpr/SandlerHZZC18}, with 1M trainable parameters and 439M FLOPs. SqueezeFaceNet \cite{DBLP:conf/ssci/YuL19}, VarGFaceNet \cite{DBLP:conf/iccvw/YanZXZWS19}, MixFaceNet \cite{DBLP:conf/icb/BoutrosDFKK21}, ShuffleFaceNet \cite{DBLP:conf/iccvw/Martinez-DiazLV19}, and GhostFaceNets \cite{ghostfacenets} adapt SqueezeNet, VarGNet, MixNet, and ShuffleNet \cite{DBLP:conf/eccv/MaZZS18} as their FR model architecture, respectively. The lightweight architectures find a trade-off between computational complexity and accuracy while keeping their computational complexity below 1G FLOPs.

To reduce the memory footprint together with the computational complexity, model quantization is also used \cite{DBLP:conf/icpr/BoutrosDK22, DBLP:journals/corr/abs-2103-13630}. This is achieved by replacing the floating point parameters of the DNN, which are often represented with 32 or 16-bit, with low-bit parameter representations, e.g. 8 or 4-bit signed integer \cite{DBLP:journals/corr/abs-2103-13630}. The lower number of bits required per parameter reduces the memory footprint of the model and reduces the computational cost, as integer arithmetic is faster than floating-point \cite{DBLP:journals/corr/abs-1806-08342}. 

In order to evaluate lightweight and efficient DNN for FR and to promote further developments, various competitions and techniques were held and developed \cite{DBLP:journals/corr/abs-2304-13409}. The Masked Face Recognition Competition at IJCB 2021 \cite{DBLP:conf/icb/BoutrosDKRKRKFZ21} has taken up the challenge of masked face recognition raised by the COVID-19 pandemic. In order to meet the emerging need for efficient FR models that can handle masked faces, a dedicated dataset was created for evaluation. The submitted solutions are ranked by their recognition performance and by the computational complexity of the proposed solution. A similar goal was conducted by the Masked Face Recognition Challenge at ICCV 2021 \cite{DBLP:conf/iccvw/DengGAZZ21}, with large-scale, private test datasets.
An extensive competition dedicated to lightweight face recognition was conducted by Deng et al. \cite{DBLP:conf/iccvw/DengGZDLS19} which further indicated a great need for efficient FR models. A dedicated lightweight face recognition model track evaluated FR models with up to 1G FLOPs. While the track introduced a limit on FLOPs, the submissions were only evaluated based on their achieved recognition performance with no further focus on the computational complexity or model size. Furthermore, the recent advancements in lightweight face recognition are not covered by this competition.

The aim of this competition is to evaluate the latest and state-of-the-art approaches for efficient and lightweight face recognition and to motivate the development of novel techniques. Unlike in previous competitions, the result of this competition will be a ranking of the submitted methods not only based on the achieved recognition performance but also with a focus on the size of the models according to the number of parameters as well as the computational effort (FLOPs) associated with the inference. The evaluation protocol published for the competition and the metrics used can be utilized in future publications as a benchmark and comparison, so that the comparability of individual methodologies can be further enhanced, being one significant impact of this competition.
The competition is organized with two different tracks. The first track includes models with up to $2$M parameters, and the second track includes models with 2-5M parameters. The competition received $17$ valid submissions from $6$ different teams. The first track received $8$, and the second track with larger models received $9$ submissions, respectively. All submissions are evaluated on common FR benchmarks introduced in Section \ref{sec:eval_datasets} and ranked based on their achieved recognition accuracy, number of parameters, and FLOPs. This paper summarizes the results of the efficient face recognition competition with a detailed presentation of the achieved recognition rates in combination with the model size and the computational complexity.

In the next section, we introduce the evaluation datasets, evaluation criteria, and the participating teams. Section \ref{sec:submitted_solutions} gives a short overview of the submitted solutions per team. Section \ref{sec:results_and_analysis} covers the achieved results. In Section \ref{sec:discussion} the methods used by the participants are discussed. Likewise, a possible outlook for currently unused approaches is given. Section \ref{sec:conclusion} ends the paper with a final general conclusion.

\section{Evaluation Setup and Participants}
\vspace{-2mm}
\subsection{Evaluation datasets}
\vspace{-2mm}
\label{sec:eval_datasets}
The submitted solutions are evaluated on a common, but diverse set of benchmarks. The solutions are ranked on the cross-pose datasets Celebrities in Frontal-Profile in the Wild (CFP-FP) \cite{cfp-fp} and Cross-Pose LFW (CPLFW) \cite{CPLFWTech}, the cross-age datasets AgeDB-30 \cite{agedb} (30 years age gap) and Cross-age LFW (CALFW) dataset \cite{CALFW}, as well as the unconstrained verification dataset Labeled Faces in the Wild (LFW) \cite{LFWTech} and the video-based verification dataset IARPA Janus Benchmark–C (IJB-C) \cite{ijbc}.

Further evaluation is performed on additional datasets. Ethnicity bias is evaluated on the Racial Faces in the Wild (RFW) \cite{rfw} dataset, which consists of Asian, African, Caucasian, and Indian subsets. Performance on cross-resolution and low-resolution images is evaluated with the verification-based Cross-Quality LFW (XQLFW) \cite{XQLFW} dataset and the identification-based TinyFace \cite{DBLP:conf/accv/ChengZG18} dataset (average 20$\times$16 pixels).

\vspace{-2mm}
\subsection{Evaluation Criteria}
\label{sec:eval_criteria}
\vspace{-2mm}
The competition is divided into two tracks based on the number of parameters of the submitted DNN. 
Teams can submit any number of solutions. The first track covers very compact networks with up to 2 million parameters (labeled $<2$ MP). The second track covers larger models from $2$ to $5$ million parameters (2-5 MP). The submitted solutions are evaluated and ranked per track. In each track, a submission is evaluated based on the achieved verification performance as well as based on the computational complexity and the memory footprint.

We follow the evaluation metrics defined in the used benchmarks. For LFW, CPLFW, CALFW, CFP-FP, XQLFW, ethnicity subsets of RFW and AgeDB30, the accuracy metric is used. For IJB-C, the true acceptance rate (TAR) at a false acceptance rate (FAR) of $10^{-4}$ is used, noted as TAR at FAR$=10^{-4}$. For both accuracy and TAR a higher value indicates better performance in comparison to a lower value.
The identification dataset TinyFace uses rank 1 and rank 5 as metrics.

To rank the individual solutions, a point system is used, the Borda count. For a considered category, the solutions are ranked by their performance, while the best-performing solution is ranked first. All $n$ solutions are assigned points based on their ranking. The first-placed solution is assigned $n-1$ points, the second-placed solution $n-2$, and the last-placed solution $0$ points, respectively.

To evaluate the recognition performance, the verification accuracy for the datasets CPLFW, CFP-FP, CALFW, AgeDB30, and LFW as well as the TAR of the IJB-C dataset are considered. The Borda counts and rankings are evaluated based on the achieved performance, and the final rank for the database is reported. A ranking over all considered benchmarks is computed based on the sum of the Borda counts of each dataset and displayed as a final database ranking.

To evaluate the deployability of a solution we consider the compactness of a model (represented by the number of parameters), the memory footprint (represented by the model size in MB), and the computational complexity (represented by M FLOPs). For all three categories, a lower value indicates better deployability in comparison to a model with higher values. The teams are asked to report the metrics (number of parameters, model size, FLOPs) for their submissions and can be asked to validate these metrics. A ranking is computed for FLOPs and model sizes using Borda count.

The final team ranking of a track is based on a weighted Borda count that uses (a) the normalized Borda count of the evaluated benchmarks (weighted $70\%$), (b) the Borda count of the FLOPs metrics (weighted $15\%$), and (c) the Borda count of the model size (weighted $15\%$). The solution with the highest final Borda count is ranked first and the solution with the lowest final Borda count is ranked last, respectively.

For example, a solution achieved Borda count $30$ summed over all verification benchmarks, Borda count $6$ in the FLOPs category and Borda count $8$ in the model size category. The normalized Borda count for the evaluated benchmarks is calculated by $\frac{30}{6}=5$. The final Borda count is calculated by $0.7\cdot5 + 0.15\cdot6+ 0.15\cdot8 = 5.6$.

\vspace{-2mm}
\subsection{Submission and Evaluation Process}
Each of the teams was asked to submit their solutions as a Linux console application or Python code with the required packages listed. The applications should accept as a parameter a text file with a list of images and an output path. The image list contains paths to aligned images for which a template is extracted and saved in the output path. The competition organizers provide pre-aligned images following \cite{arcface}. If a participant uses a different alignment method, a dedicated alignment script is submitted by the team. The script is provided with image paths, bounding boxes, and landmarks. The bounding boxes and landmarks are detected using pre-trained Multitask Cascaded Convolutional Networks (MTCNN) \cite{DBLP:journals/spl/ZhangZLQ16}. The extracted features are used with the respective protocol of the evaluation dataset. The participants are free to choose their training data. However, these databases should be publicly accessible and the authors should have a license (if required) to use these databases (when required by the data creators). The participants take full responsibility for ensuring the proper legal and ethical use of the data.
\vspace{-2mm}
\subsection{Competition Participants}
The competition aimed at attracting teams from various research institutes with high geographic and activity variation. The call for participation was shared on the International Joint Conference on Biometrics (IJCB 2023) website, on the dedicated competition website, on mailing lists, and on social media. The call has attracted 7 registered teams. Of these, 6 teams submitted at least one solution. Each of the teams is from a different country.
Each team was allowed to submit any number of submissions for both tracks. A total of 17 different submissions were submitted by the teams. For the 2-5 MP track $9$ solutions were submitted, and $8$ solutions for the $<2$ MP track.
%
%
\begin{table*}
\centering
\resizebox{\textwidth}{!}{%
\begin{tabular}{|l|l|l|} 
\hline
\multicolumn{1}{|c|}{\textbf{Solution}}                                                                                             & \multicolumn{1}{c|}{\textbf{Team members}}                                                                                         & \multicolumn{1}{c|}{\textbf{Affiliations}}                                                                                                                                                                                  \\ 
\hhline{|===|}
GhostFaceNetV1-1 KU,~GhostFaceNetV1-2 KU                                                                                            & \begin{tabular}[c]{@{}l@{}}Mohamad Alansari, Oussama Abdul Hay,\\Sara Alansari, Sajid Javed, Naoufel Werghi\end{tabular}           & \begin{tabular}[c]{@{}l@{}}Department of Electrical and Computer Engineering, Khalifa University, \\Abu Dhabi, United Arab Emirates\end{tabular}                                                                            \\ 
\hline
ShuffleNetv2x0.5,~ShuffleNetv2x1.5,~ShuffleNetv2x2.0                                                                                & Klemen Grm, Vitomir Štruc                                                                                                          & \begin{tabular}[c]{@{}l@{}}Laboratory for Machine Intelligence, Faculty of Electrical Engineering,\\University of Ljubljana, Slovenia, EU\end{tabular}                                                                      \\ 
\hline
SQ-HH,~MB2-HH                                                                                                                       & Fernando Alonso-Fernandez, Kevin Hernandez Diaz, Josef Bigun                                                                       & Halmstad University, Sweden                                                                                                                                                                                                 \\ 
\hline
\begin{tabular}[c]{@{}l@{}}Idiap EdgeFace-XS($\gamma$=0.6),Idiap EdgeFace-XXS-Q,\\Idiap EdgeFace-S($\gamma$=0.5),Idiap EdgeFace-XS-Q\end{tabular} & \begin{tabular}[c]{@{}l@{}}Anjith George, Christophe Ecabert,\\Hatef Otroshi Shahreza, Ketan Kotwal, Sébastien Marcel\end{tabular} & \begin{tabular}[c]{@{}l@{}}Idiap Research Institute, Martigny, Switzerland.\\École Polytechnique Fédérale de Lausanne (EPFL), Lausanne, Switzerland. \\Université de Lausanne (UNIL), Lausanne, Switzerland.\end{tabular}  \\
\hline
MobileNet$_{\text{V2-visteam}}$,~EfficientNet$_{\text{b0-visteam}}$                                                                                   & Iurii Medvedev, Bo Jin, Diogo Nunes                                                                                                & Institute of Systems and Robotics, University of Coimbra, Coimbra, Portugal                                                                                                                                                 \\ 
\hline
SAM-MFaceNet eHWS,~Modified-MobileFaceNet                                                                                           & Ahmad Hassanpour, Pankaj Khatiwada, Aafan Ahmad Toor, Bian Yang                                                                    & \begin{tabular}[c]{@{}l@{}}eHealth and Welfare Security Group, Department of Information Security and\\Communication Technology, Norwegian University of Science and Technology, Norway\end{tabular}                        \\
\hline
\end{tabular}
}
\vspace{1mm}
\caption{A summary of the submitted solutions, participant team members, and affiliations.}
\label{tab:submissions_overview}

\end{table*}

\begin{table}
\centering
\resizebox{\linewidth}{!}{%
\begin{tabular}{|c|c|c|c|c|c|} 
\hline
\textbf{Solution}               & \textbf{Architecture} & \textbf{Loss function} & \textbf{Optimizer} & \textbf{Training datasets} & \textbf{Feature size}  \\ 
\hline \hline
GhostFaceNetV1-1 KU             & GhostNetV1            & ArcFace                & SGD                & MS1MV3                     & 512                    \\ 
\hline
GhostFaceNetV1-2 KU             & GhostNetV1            & ArcFace                & SGD                & MS1MV3                     & 512                    \\ 
\hline
ShuffleNetv2x0.5                & ShuffleNet v2         & Cross-entropy          & AdamW              & VGGFace2                   & 128                    \\ 
\hline
ShuffleNetv2x1.5                & ShuffleNetV2          & Cross-entropy          & AdamW              & VGGFace2                   & 345                    \\ 
\hline
ShuffleNetv2x2.0                & ShuffleNetV2          & Cross-entropy          & AdamW              & VGGFace2                   & 1666                   \\ 
\hline
SQ-HH                           & Squeezenet            & Softmax                & SGD                & MS1M, VGG2                 & 1000                   \\ 
\hline
MB2-HH                          & MobileNetV2           & Softmax                & SGD                & MS1M, VGG2                 & 1280                   \\ 
\hline
Idiap EdgeFace-XS($\gamma$=0.6) & EdgeNeXt              & CosFace                & AdamW              & WebFace12M                 & 512                    \\ 
\hline
Idiap EdgeFace-XXS-Q            & EdgeNeXt              & CosFace                & AdamW              & WebFace4M                  & 512                    \\ 
\hline
Idiap EdgeFace-S($\gamma$=0.5)  & EdgeNeXt              & CosFace                & AdamW              & WebFace12M                 & 512                    \\ 
\hline
Idiap EdgeFace-XS-Q             & EdgeNeXt              & CosFace                & AdamW              & WebFace4M                  & 512                    \\ 
\hline
MobileNet$_{\text{V2-visteam}}$          & MobileNetV2           & ArcFace                & SGD                & WebFace42M                 & 512                    \\ 
\hline
EfficientNet$_{\text{b0-visteam}}$       & EfficientNet\_b0      & ArcFace                & SGD                & WebFace42M                 & 512                    \\ 
\hline
SAM-MFaceNet eHWS               & MobileFaceNet         & MagFace                & SAM                & WebFace42M                 & 512                    \\ 
\hline
Modified-MobileFaceNet          & MobileFaceNet         & MagFace                & SAM                & WebFace42M                 & 512                    \\
\hline
\end{tabular}
}

\caption{Details of the submitted solutions including the backbone model architecture, the used loss function, optimizer and datasets during training, and the feature size used. All solutions use $112\times112$ image size, except SQ-HH and MB2-HH that utilize images sized $113\times113$.}
\label{tab:submission_detail_overview}
\vspace{-4mm}
\end{table}

\section{Submitted Solutions and Baselines}
\label{sec:submitted_solutions}
An overview of the participating teams with their respective team members, affiliation, and the submitted solution is shown in Table \ref{tab:submissions_overview}. Detailed information for each solution is shown in Table \ref{tab:submission_detail_overview}.

\subsection{GhostFaceNets}
Mixed-precision FP32 GhostNetV1 (GhostFaceNetV1-1 KU) and GhostNetV2 (GhostFaceNetV1-2 KU) architecture \cite{ghostnet} were trained from scratch using the ArcFace loss function \cite{arcface} on the MS1MV3 dataset \cite{arcface}. The Stochastic Gradient Descent (SGD) optimizer with a momentum of 0.9 and weight decay of 5e-4 was employed, along with a cosine learning scheduler. The learning rate base was set to 0.1, the learning rate decay was set to 0.5, the learning rate decay steps were set to 44, and the minimum learning rate was set to 1e-5. During the training process, a batch size of 512 was used, and the training epoch was set to 50. To mitigate overfitting, \(l_2\) regularization with \(l_2 = 1/2\) was applied to the model's output layer. For verification experiments, cosine distance was utilized. For more detailed information about the models, refer to the publication on GhostFaceNets \cite{ghostfacenets}.

\subsection{LMI\_ShuffleNetv2}
The LMI\_ShuffleNetv2 submitted three models based on ShuffleNetV2 architecture \cite{ma2018shufflenet}.
All submitted models, ShuffleNetv2x0.5, ShuffleNetv2x1.5, and ShuffleNetv2x2.0 utilized full-precision FP32.  These architectures were varied in size by scaling the width of every convolutional layer of ShuffleNetV2 by a factor of 0.5, 1.5, and 2,  respectively. The embeddings are derived by global average pooling over the response of the final convolutional layer. Furthermore, the dimensionality of the embedding layer was decreased to 128 by reducing the number of filters in the final convolutional layer. To train the model, a linear classifier was used that projects the embedding into an 8631-dimensional logit vector. As a training set, the VGGFace2 dataset is used \cite{cao2018vggface2}. All images are aligned using the code provided by the organizers. This results in a dataset of 2.76M $112\times112$ images of 8631 subjects. This dataset is used to train the model with a cross-entropy loss, using the AdamW optimizer \cite{loshchilov2017decoupled} and a batch size of 64, and cosine learning rate annealing from $3\times10^{-4}$ towards $10^{-6}$.  At test time, the classifier and the activation function over the embedding were removed. After training,  the pointwise mean and standard deviation vectors of the embeddings over the training dataset are recorded and used to normalize the embeddings at test time, as $ \mathbf{v}_{emb} = \left( \mathbf{v}_{emb} - \mathbf{v}_{\mu} \right) / \mathbf{v}_{\sigma}$.

\subsection{SQ-HH}
HH-MB2 and HH-SQ employ a MobileNetv2 \cite{DBLP:conf/cvpr/SandlerHZZC18} and SqueezeNet \cite{DBLP:journals/corr/IandolaMAHDK16_squeezenet} backbone, respectively. ImageNet pre-trained networks used as initial weights
SQ-HH has also added batch normalization between convolutions and ReLU layers of SqueezeNet, which are not included in their original implementation. Both CNNs have been modified to employ an input size of 113$\times$113$\times$3 by changing the stride of the first convolutional layer from 2 to 1. This allows to keep the rest of the network unchanged and to reuse ImageNet parameters as starting model. Then, the networks undergo a double fine-tuning, first over MS1M-RetinaFace cleaned set \cite{DBLP:conf/eccv/GuoZHHG16} (35k subjects/3.16M images, only subjects with more than 70 images), and then over VGGFace2 \cite{cao2018vggface2} (9k subjects/3.31M images). Although VGG2 contains fewer identities, it has more intra-class diversity due to more images per subject. Due to this fact, the double fine-tuning strategy employed has shown increased performance compared to training the models only with one database, especially if it has few images per identity \cite{Alonso20SqueezeFacePoseNet, cao2018vggface2}.
The networks are trained for biometric identification using the soft-max function and ImageNet as initialization. The optimizer is SGDM (mini-batch=128, learning rate=0.01, 0.005, 0.001, and 0.0001, decreased when the validation loss plateaus). Two percent of images per subject in the training set are set aside for validation. The proposed models are trained with Matlab r2022b and use the ImageNet pre-trained model that comes with such a release.
After training, feature vectors for biometric verification are extracted from the layer adjacent to the classification layer (i.e., the Global Average Pooling). This corresponds to a descriptor of 1280 (HH-MB2) and 1000 (HH-SQ) elements.

\subsection{Idiap EdgeFace}
Idiap EdgeFace submitted four solutions, Idiap EdgeFace-XS($\gamma$=0.6), Idiap EdgeFace-XXS-Q, Idiap EdgeFace-S($\gamma$=0.5), and Idiap EdgeFace-XS-Q. These models adopt the EdgeNeXt \cite{DBLP:conf/eccv/MaazSCKZAK22} architecture to create lightweight face recognition models. Specifically, the S, XS, and XXS variants of the architecture were adopted by introducing a 512-D classification head and introducing low-rank linear layers, where $\gamma$ is the ratio of the rank. The details of these models can be found in the publication \cite{george2023edgeface}. Idiap EdgeFace-S($\gamma$=0.5) and Idiap EdgeFace-XS($\gamma$=0.6) utilize full-precision FP32. Idiap EdgeFace-XS-Q and EdgeFace-XXS-Q utilize quantization to 8-bit integers.
CosFace \cite{DBLP:conf/cvpr/WangWZJGZL018_cosface} is used as the loss function and trained the network from scratch using a polynomial decaying training schedule with restarts for 200 epochs along with learning rate warm-up for 2 epochs. The models were trained with 4 and 8 Nvidia RTX 3090 (24GB) GPUs using a distributed training strategy. The submitted models were trained using PyTorch with AdamW optimizer \cite{loshchilov2017decoupled} with a weight decay of $5\times10^{-2}$, and a learning rate of $3\times10^{-3}$. The models were trained on WebFace4M and the WebFace12M dataset \cite{WebFace260M}. The batch size used was 512. Data augmentations such as random horizontal flips, random blur, grayscale, and resizing were used to improve the robustness of the model.

\subsection{Visteam}
The team submitted two half-precision models EfficientNet$_{\text{b0-visteam}}$ and MobileNet$_{\text{V2-visteam}}$ based on EfficientNet$_{\text{B0}}$  \cite{mobilenet} and MobileNetV2 \cite{DBLP:conf/cvpr/SandlerHZZC18} architecture. Both models were trained from scratch using ArcFace \cite{arcface} loss function.  
The deep embedding during the training process is supervised by a large in-house model. Training is performed on the WebFace42M dataset \cite{WebFace260M}. To reduce the number of classes in the original dataset, a subset is created with identities that contain at least $50$ images. The Stochastic Gradient Descent (SGD) optimizer with a momentum of 0.9 and weight decay of 1e-4. 
The learning rate linearly decays from 0.01 to 0.0001 for 15 epochs. The batch size in the training phase was 128. The conventional data augmentation techniques horizontal flipping, random gamma adjustment, RGB shifting, color jittering, and occlusion were used to avoid overfitting.
After training the full-precision network was quantized to half-precision floating point.

\subsection{SAM-MFaceNet eHWS}
The team submitted two models, SAM-MFaceNet eHWS and Modified-MobileFaceNet.
In SAM-MFaceNet eHWS, a pre-trained full-precision FP32 MobileFaceNet architecture was trained using the MagFace loss function on a portion of the WebFace42M dataset including 100K identities and 4.2M images. 
In Modified-MobileFaceNet, MobileFaceNet architecture was modified by doubling the kernels of the first three CNN layers. It was trained from scratch using the MagFace \cite{DBLP:conf/cvpr/MengZH021_Magface} loss function on a portion of WebFace42M \cite{WebFace260M} dataset including 100K identities.
For both models, the sharpness-aware minimization (SAM) \cite{foret2020sharpness}
optimizer with $\rho = 0.05$ and an exponential learning scheduler with a gamma of 0.998 was applied in the training processes. The initial learning rate for training the ResNet models was set to 0.1. The batch size in the training phase was 256 and the training epoch was set to 100. 

\subsection{Baselines}
Baselines have been selected to compare the submitted approaches with state-of-the-art FR model performance. A diverse selection of previously published approaches for efficient FR is given to allow a comparison with a range of existing methods (e.g. large ResNet models \cite{DBLP:conf/cvpr/BoutrosDKK22, arcface, DBLP:conf/cvpr/HeZRS16}, winner of the ICCV 2019 Lightweight Face Recognition challenge VarGFaceNet \cite{DBLP:conf/iccvw/YanZXZWS19}, PocketNet \cite{DBLP:journals/access/BoutrosSKDKK22}, MobileFaceNet \cite{DBLP:conf/ccbr/ChenLGH18}).

\section{Results and Analysis}
\label{sec:results_and_analysis}
This section first introduces and compares the results of the baselines and the proposed solutions on the benchmarks used to determine the final ranking of the submissions. This analysis is followed by an additional in-depth evaluation of bias, cross-quality, and a large-scale recognition benchmark.

\subsection{Ranking benchmarks}
Table \ref{tbl:overall_result_with_borda_count} shows an overview of the baseline models and the submitted solution on the employed benchmarks including rankings of the participants' solutions. The ranking is performed based on the evaluation criteria described in Section \ref{sec:eval_criteria}.

The largest baseline models use ResNet-100 as a backbone architecture. The network is trained once with the ElasticFace-Cos+ and ArcFace loss functions. On average, both models achieve the highest accuracies on the selected benchmarks. VarGFaceNet, the winner of the ICCV 2019 Lightweight Face Recognition Challenge \cite{DBLP:conf/iccvw/DengGZDLS19} has only slightly lower accuracies than the ResNet-100 model, especially on the cross-pose dataset CPLFW, but achieves these comparable results with only $8\%$ of the parameters of the large ResNet-100 model. And thus VarGFaceNets requires only 20MB of memory footprint instead of the 261.22MB of the ResNet-100 models. 
The ShuffleMixFaceNets \cite{DBLP:conf/icb/BoutrosDFKK21} models have in comparison to ResNet-100 models a slightly lower verification accuracy on the AgeDB30, LFW, and IJB-C benchmarks. ShuffleMixFaceNet-XS has a $4$ percentage point reduction in accuracy on IJB-C in comparison to VarGFaceNet. But it uses only $161.9$ FLOPs instead of $1022$ FLOPs of VarGFaceNet.
ShuffleFaceNet 1.5x \cite{DBLP:conf/iccvw/Martinez-DiazLV19} achieves similar performance to VarGFaceNet but requires only half of the number of FLOPs. It has 2.6M parameters in comparison to VarGFaceNets 5.0M and therefore requires a memory footprint of 10.5MB.
The PocketNet \cite{DBLP:journals/access/BoutrosSKDKK22} models achieve a small increase in verification accuracies (around $1$ percentage point) in comparison to ShuffleMixFaceNet-XS, but the number of FLOPs of these models has  $6$-fold increase when compared to ShuffleMixFaceNet-XS.

From the results of the experiment of the models with 2-5 million parameters, the following conclusions can be drawn:
 \vspace{-2mm}
\begin{itemize}
    \item The overall top-ranked solution is GhostFaceNetV1-1 KU (rank 1), followed by GhostFaceNetV1-2 KU (rank 2) and Idiap EdgeFace-S($\gamma$=0.5) (rank 3). \vspace{-2mm}
    \item Highest ranked solution in terms of verification accuracies over the evaluated benchmarks is Idiap EdgeFace-S($\gamma$=0.5). It also achieved the highest TAR at FAR=$10^{-4}$ on the large IJB-C benchmark for all submitted solutions in this category. The ROC plots for models with 2-5 million parameters are shown in Figure \ref{fig:roc_large_ijbc}.
     \vspace{-2mm}
    \item GhostFaceNetV1-2 KU has the lowest number of FLOPs in the model category, and its number of FLOPs is lower than any of the baseline models.
     \vspace{-2mm}
    \item The lowest model size in MB is achieved by Idiap EdgeFace-XS-Q. With 2.24M parameters that are quantized to $8$-bit integers, the model results in 2.99MB of memory footprint. In comparison, the baseline with the lowest number of parameters (ShuffleMixFaceNet-XS) has a model size o 4.16MB, as $32$-bit floating-point numbers are used for each of the 1.04M parameters.
     \vspace{-2mm}
    \item No submission of the category achieves high ranks in both cross-pose and cross-age benchmarks simultaneously.
     \vspace{-2mm}
    \item The model size and therefore the number of parameters of the submitted solution is in a similar range to that of the baseline models.
     \vspace{-2mm}
    \item The highest ranked solution GhostFaceNetV1-1 KU has at most a $3.5$ percentage-points difference to the best-performing baseline (ResNet-100 ElasticFace-Cos+), but requires only $0.89\%$ of the number of FLOPs to the ResNet-100 model. 
    When computational complexity and verification accuracies are considered, it outperforms the respective baseline.
     \vspace{-2mm}
\end{itemize}

For solutions with $<2$ million parameters, the following can be observed:
 \vspace{-2mm}
\begin{itemize}
    \item Idiap EdgeFace-XS($\gamma$=0.6) is the overall best-performing model, SAM-MFaceNet eHWS V1 is ranked 2, SAM-MFaceNet eHWS V2 is ranked 3, respectively.
    \item Idiap EdgeFace-XS($\gamma$=0.6) also ranks first when considering the verification accuracy, including the first rank on IJB-C. The ROC curves for all submissions with $<2$ million parameters are shown in Figure \ref{fig:roc_small_ijbc}. 
    \item The smallest and most efficient model is ShuffleNetv2x0.5 (rank 7), with 0.17M parameters resulting in $0.77$MB of model size and 17.14 FLOPs.
    \item Idiap EdgeFace-XS($\gamma$=0.6) has at most a $4.5$ percentage-points difference to the best-performing baseline (ResNet-100 ElasticFace-Cos+) but requires only $0.63\%$ of the number of FLOPs to the ResNet-100 model.
\end{itemize}

Almost all of the submitted models, either with $<2$M parameters or 2-5M parameters, follow \cite{arcface} and \cite{DBLP:conf/cvpr/WangWZJGZL018_cosface} respectively with the alignment procedure, but also with the embedding size. All top-ranking models use a margin-penalty-based loss function like ArcFace \cite{arcface}, CosFace \cite{DBLP:conf/cvpr/WangWZJGZL018_cosface}, or MagFace \cite{DBLP:conf/cvpr/MengZH021_Magface}. The majority of submissions use either SGD or AdamW as an optimizer. Only three different base datasets have been selected for training, which differ in their respective version primarily in their number of training samples and different processing steps. The data sets used are different versions of MS-Celeb-1M \cite{DBLP:conf/eccv/GuoZHHG16}, VGGFace2 \cite{cao2018vggface2}, and WebFace260M \cite{WebFace260M}. The submissions use similar network architectures: GhostNet \cite{DBLP:conf/cvpr/HanW0GXX20_GhostNet}, ShuffleNet \cite{ma2018shufflenet}, EdgeNext \cite{DBLP:conf/eccv/MaazSCKZAK22}, and MobileNetV2 \cite{DBLP:conf/cvpr/SandlerHZZC18} are the most used backbone architectures.
These models are established architectures that focus on efficient execution.

\subsection{Further Benchmarks}
We extend our evaluation beyond the competition ranking by evaluating bias (RFW), cross-quality (XQLFW), low-resolution (TinyFace), and large-scale recognition (IJB-C) benchmarks evaluated at different operation thresholds. The results of the additional evaluation are shown in Table \ref{tbl:evaluation}

To evaluate the bias of each submitted solution, we follow \cite{DBLP:conf/cvpr/WangD20} and calculate the verification accuracies for Asian, African, Caucasian, and Indian identities of the RFW dataset. The average and standard deviation of the accuracies, as well as the skewed error ratio (SER), are computed. A higher SER indicates that a model inhibits a higher bias than a model with a lower SER value. The results show that the top-ranking models (GhostFaceNetV1-1 KU, Idiap EdgeFace-XS($\gamma$=0.6)) are inhibiting a larger bias when compared to other submitted models. Furthermore, for all models, it holds that Caucasian achieve the highest verification accuracies.

On XQLFW, Idiap EdgeFace-S($\gamma$=0.5) (2-5M parameters group) has achieved the highest verification accuracy, followed by Idiap EdgeFace-XS($\gamma$=0.6) ($<2$M parameters) and Idiap EdgeFace-XS-Q (2-5M parameters). No large deviation of verification accuracy can be observed between models with 2-5M parameters and $<2$M parameters.

On the LR benchmark TinyFace, the highest rank 1 and rank 5 accuracy was achieved by Modified-MobileFaceNet V1, followed by Modified-MobileFaceNet V2 and SAM-MFaceNet eHWS V1.

\begin{figure}
\centering
\begin{subfigure}[h!]{0.65\linewidth}
\includegraphics[width=\linewidth]{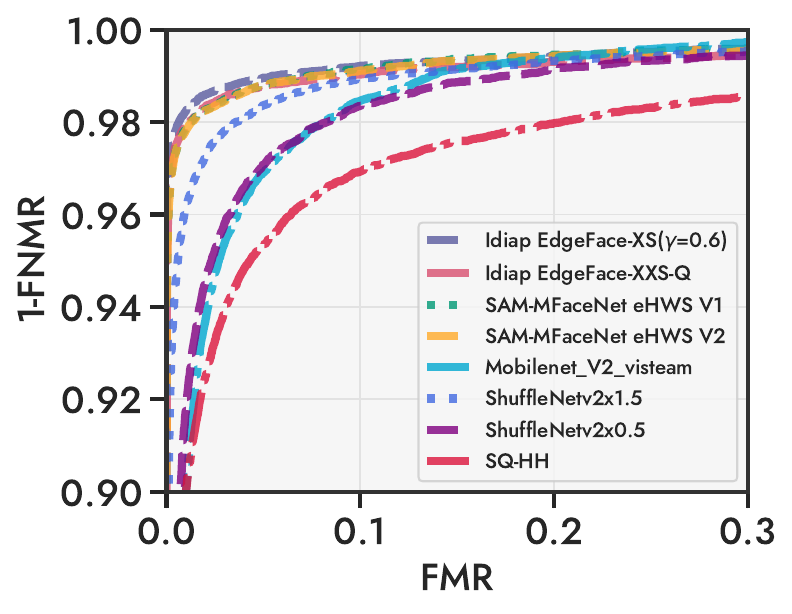}
\caption{$<2$M parameters models}
\label{fig:roc_small_ijbc}
\end{subfigure}
\hfill
\begin{subfigure}[h!]{0.65\linewidth}
\includegraphics[width=\linewidth]{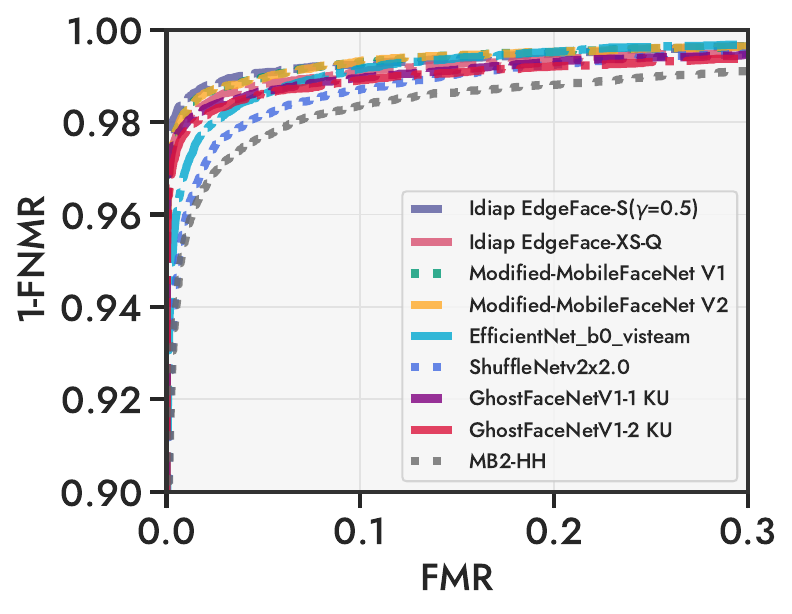}
\caption{2-5M parameters models}
\label{fig:roc_large_ijbc}
\end{subfigure}%
\caption{The ROC curve scored by models in the respective track on the IJB-C benchmark.}
\vspace{-2mm}
\end{figure}

On the large-scale verification benchmark IJB-C, all FAR thresholds of the test set are shown in Table \ref{tbl:evaluation}.
The best TAR at the lowest FAR=$10^{-6}$ is achieved by GhostFaceNetV1-1 KU with $86.38\%$. The second best TAR is achieved by GhostFaceNetV1-2 KU with $85.82\%$, respectively (both 2-5M parameters). While GhostFaceNetV1-1 KU showed a strong performance at low FAR, Idiap EdgeFace-S($\gamma$=0.5) achieved the top TAR at higher FAR. This is visible as well in the ROC plot shown in Figure \ref{fig:roc_large_ijbc}.
The top performing models in the $<2$M parameters category show a strong degradation of performance at FAR=$10^{-6}$ when compared to models in the 2-5M parameters group. At higher FAR values, the models of the small model category achieve similar TAR  as models of the large model group.

\section{Discussion}
\label{sec:discussion}
The methodologies used to develop the submitted efficient models are primarily compact network architectures. The top-performing submissions are adapting the architectures GhostNet and EgdeNeXt.
Two submissions, Idiap EdgeFace-XS-Q, and Idiap EdgeFace-XXS-Q \cite{george2023edgeface}, are using model quantization with 8-bit integer quantization, similar to \cite{DBLP:conf/icpr/BoutrosDK22, DBLP:conf/icb/KolfBKD22, DBLP:journals/ivc/KolfEB0D23}.
Techniques that were unused by the submissions include knowledge distillation, pruning and sparsity-inducing training, neural architecture search, and binary neural networks.

In knowledge distillation \cite{DBLP:journals/ijcv/GouYMT21}, the knowledge of a larger, high-performing teacher network is transferred into a smaller student network.
Pruning \cite{DBLP:journals/ijon/LiangGWSZ21} involves dropping neurons or convolutional filters, which are not critical to the high performance of the network. If computational components can be removed without significant performance loss, a much more efficient and memory-saving model can be created.
A similar approach is sparsity-induced training \cite{DBLP:journals/corr/abs-1902-09574}. There, as many weights of the network as possible are set to 0 during the training process. This creates sparse matrices that are simpler and more efficient to compute, depending on the architecture.
Neural architecture search \cite{DBLP:journals/jmlr/ElskenMH19}, as used in the baseline PocketNet \cite{DBLP:journals/access/BoutrosSKDKK22}, tries to learn by an optimization algorithm as compact and small network components as possible, which are adapted to the respective learning task and therefore achieve higher performance with less computational effort.
Instead of using float-point or integer parameters, binary neural networks \cite{yuan2023comprehensive} use binary values for the computations. This reduces the calculation effort considerably since significantly fewer logical operations are required for the calculations. Furthermore, it reduces the memory footprint of the model, as only a single bit per parameter is stored.
Due to the described different, currently unused techniques, there is considerable scope for application and development to make efficient FR models even more compact and powerful.

\section{Conclusion}
\label{sec:conclusion}
The aim of this Efficient Face Recognition Competition (EFaR) was to benchmark and evaluate current technologies for lightweight and efficient face recognition. The submissions were evaluated with respect to the achieved verification accuracies on a diverse benchmark suite, in combination with their deployability given by the number of floating-point operations and the required memory footprint. Furthermore, additional evaluation was performed on bias, cross-quality, and large-scale face recognition benchmarks. The results show that, when deployability is considered as well, top-ranking efficient face recognition solutions achieve comparable results to large face recognition models. Further, it is observed, that the majority of top-ranking solutions inhibit racial bias and achieve similar performance on XQLFW and IJB-C benchmarks. While the submissions show high-performing, efficient face recognition models, further, currently unused techniques for the design of lightweight face recognition models can be adopted.

\textbf{Acknowledgment}
This research work has been funded by the German Federal Ministry of Education and Research and the Hessian Ministry of Higher Education, Research, Science and the Arts within their joint support of the National Research Center for Applied Cybersecurity ATHENE. This work has been partially funded by the German Federal Ministry of Education and Research (BMBF) through the Software Campus Project.

\begin{table*}
\centering
\resizebox{\textwidth}{!}{%
\begin{tabular}{c|c|cc|cc||cc|cc||cc|cc|cc|cc|cc|c|cc}
\multicolumn{1}{l}{}             & \multicolumn{1}{l}{} & \multicolumn{4}{c||}{\textbf{Cross-Pose}}                                   & \multicolumn{4}{c||}{\textbf{Cross-Age}}                                         & \multicolumn{2}{l}{}               & \multicolumn{2}{l}{}                   & \multicolumn{2}{l}{}                   & \multicolumn{2}{l}{}                                     & \multicolumn{2}{l}{}                     & \multicolumn{1}{l}{} & \multicolumn{2}{l}{}                   \\
\multicolumn{1}{c}{}             &                      & \multicolumn{2}{c|}{\textbf{CPLFW}} & \multicolumn{2}{c||}{\textbf{CFP-FP}} & \multicolumn{2}{c|}{\textbf{CALFW}} & \multicolumn{2}{c||}{\textbf{AgeDB30}}    & \multicolumn{2}{c|}{\textbf{LFW}}  & \multicolumn{2}{c|}{\textbf{IJB-C}}    & \multicolumn{2}{c|}{\textbf{Accuracy}} & \multicolumn{2}{c|}{\textbf{FLOPS}}                      & \multicolumn{2}{c|}{\textbf{Model Size}} & \textbf{Params}      & \multicolumn{2}{c}{\textbf{Combined}}  \\
\textbf{Model}                   & \textbf{Category}    & \textbf{Acc. [\%]} & \textbf{Rank}  & \textbf{Acc. [\%]} & \textbf{Rank}   & \textbf{Acc. [\%]} & \textbf{Rank}  & \textbf{Acc. [\%]} & \textbf{Rank}       & \textbf{Acc. [\%]} & \textbf{Rank} & \textbf{TAR@$10^{-4}$} & \textbf{Rank} & \textbf{BC} & \textbf{Rank}            & \textbf{[M]}                        & \textbf{Rank} & \textbf{[MB]} & \textbf{Rank}            & \textbf{[M]}         & \textbf{BC} & \textbf{Rank}            \\ 
\hline
ResNet-100 ElasticFace (Cos+) \cite{DBLP:conf/cvpr/BoutrosDKK22}   & Baseline             & 93.23              & -              & 98.73              & -               & 96.18              & -              & 98.28              & -                   & 99.80              & -             & 96.65                  & -             & -           & -                        & 24211.778                                & -             & 261.22        & -                        & 65.2                 & -           & -                        \\ 
\hline
ResNet-100 ArcFace \cite{DBLP:conf/cvpr/BoutrosDKK22, arcface}              & Baseline             & 92.08              & -              & 98.27              & -               & 95.45              & -              & 98.15              & -                   & 99.82              & -             & 95.60                  & -             & -           & -                        & 24211.778                                & -             & 261.22        & -                        & 65.2                 & -           & -                        \\ 
\hline
ResNet-18 Q8-bit \cite{DBLP:conf/icpr/BoutrosDK22}                 & Baseline             & 89.48              & -              & 94.46              & -               & 95.72              & -              & 97.03              & -                   & 99.63              & -             & 93.56                  & -             & -           & -                        & \textcolor[rgb]{0.149,0.149,0.149}{1810} & -             & 24.10         & -                        & 24.0                 & -           & -                        \\ 
\hline
ResNet-18 Q6-bit \cite{DBLP:conf/icpr/BoutrosDK22}                & Baseline             & 88.37              & -              & 93.23              & -               & 95.58              & -              & 96.55              & -                   & 99.52              & -             & 93.03                  & -             & -           & -                        & \textcolor[rgb]{0.149,0.149,0.149}{1810} & -             & 18.10         & -                        & 24.0                 & -           & -                        \\ 
\hline\hline
VarGFaceNet \cite{DBLP:conf/iccvw/YanZXZWS19}                     & Baseline             & 88.55              & -              & 98.50              & -               & 95.15              & -              & 98.15              & -                   & 99.85              & -             & 94.70                  & -             & -           & -                        & 1022                                     & -             & 20.0          & -                        & 5.0                  & -           & -                        \\ 
\hline
MobileFaceNetV1 \cite{DBLP:conf/ccbr/ChenLGH18}                 & Baseline             & 87.17              & -              & 95.80              & -               & 94.47              & -              & 96.40              & -                   & 99.40              & -             & 93.90                  & -             & -           & -                        & 1100                                     & -             & 13.6          & -                        & 3.4                  & -           & -                        \\ 
\hline
ShuffleMixFaceNet-M  \cite{DBLP:conf/icb/BoutrosDFKK21}            & Baseline             & 89.97              & -              & 94.96              & -               & 95.75              & -              & 96.98              & -                   & 99.60              & -             & 91.47                  & -             & -           & -                        & 626.1                                    & -             & 15.8          & -                        & 3.95                 & -           & -                        \\ 
\hline
ShuffleMixFaceNet-S \cite{DBLP:conf/icb/BoutrosDFKK21}             & Baseline             & 89.85              & -              & 94.10              & -               & 95.67              & -              & 97.05              & -                   & 99.58              & -             & 93.08                  & -             & -           & -                        & 451.7                                    & -             & 12.28         & -                        & 3.07                 & -           & -                        \\ 
\hline
ShuffleMixFaceNet-XS  \cite{DBLP:conf/icb/BoutrosDFKK21}           & Baseline             & 86.93              & -              & 91.25              & -               & 94.93              & -              & 95.61              & -                   & 99.53              & -             & 90.43                  & -             & -           & -                        & 161.9                                    & -             & 4.16          & -                        & 1.04                 & -           & -                        \\ 
\hline
ShuffleFaceNet 1.5x \cite{DBLP:conf/iccvw/Martinez-DiazLV19}             & Baseline             & 88.50              & -              & 97.26              & -               & 95.05              & -              & 97.32              & -                   & 99.67              & -             & 94.30                  & -             & -           & -                        & 577.5                                    & -             & 10.5          & -                        & 2.6                  & -           & -                        \\ 
\hline
MobileFaceNet \cite{DBLP:conf/ccbr/ChenLGH18}                   & Baseline             & 89.22              & -              & 96.90              & -               & 95.20              & -              & 97.60              & -                   & 99.70              & -             & 94.70                  & -             & -           & -                        & 933                                      & -             & 4.50          & -                        & 2.0                  & -           & -                        \\ 
\hline
EfficientNet$_\text{b0-visteam}$ & 2-5 MP               & 87.58              & 9              & 91.19              & 9               & 93.35              & 7              & 90.45              & 7                   & 99.15              & 8             & 85.04                  & 7             & 7           & 8                        & 212.50                                   & \textit{3}    & 9.18          & 7                        & 4.60                 & 1.87        & 8                        \\ 
\hline
GhostFaceNetV1-1 KU \cite{ghostfacenets}             & 2-5 MP               & 91.70              & 4              & 95.00              & 5               & 95.77              & \textbf{1}     & 97.20              & \textbf{1}          & 99.62              & \textit{3}    & 94.93                  & \uline{2}     & 37          & \uline{2}                & 215.65                                   & 4             & 8.17          & 4                        & 4.09                 & 5.52        & \textbf{1}               \\ 
\hline
GhostFaceNetV1-2 KU \cite{ghostfacenets}             & 2-5 MP               & 90.03              & 7              & 93.30              & 7               & 95.72              & \uline{2}      & 97.08              & \uline{2}           & 99.72              & \uline{2}     & 94.06                  & 4             & 29          & 5                        & 60.29                                    & \textbf{1}    & 8.07          & \uline{2}                & 4.06                 & 5.33        & \uline{2}                \\ 
\hline
Idiap EdgeFace-S($\gamma$=0.5) \cite{george2023edgeface}  & 2-5 MP               & 92.22              & \textit{3}     & 95.67              & \textit{3}      & 95.62              & \textit{3}     & 96.98              & \textit{3}          & 99.78              & \textbf{1}    & 95.63                  & \textbf{1}    & 39          & \textbf{1}               & 306.11                                   & 5             & 14.69         & 8                        & 3.65                 & 5.15        & \textit{3}               \\ 
\hline
Idiap EdgeFace-XS-Q  \cite{george2023edgeface}            & 2-5 MP               & 90.92              & 5              & 94.26              & 6               & 95.03              & 6              & 95.22              & 6                   & 99.50              & 6             & 94.40                  & \textit{3}    & 22          & 6                        & 196.91                                   & \uline{2}     & 2.99          & \textbf{1}               & 2.24                 & 4.52        & 4                        \\ 
\hline
MB2-HH                           & 2-5 MP               & 90.65              & 6              & 95.13              & 4               & 91.43              & 8              & 90.08              & 8                   & 99.32              & 7             & 79.86                  & 9             & 12          & 7                        & 722.59                                   & 9             & 8.15          & \textit{3}               & 2.20                 & 2.15        & 7                        \\ 
\hline
Modified-MobileFaceNet V1        & 2-5 MP               & 92.42              & \textbf{1}     & 95.97              & \uline{2}       & 95.15              & 4              & 95.77              & 5                   & 99.52              & 5             & 93.99                  & 5             & 31          & 4                        & 456.89                                   & 7             & 8.4           & 5                        & 2.10                 & 4.22        & 6                        \\ 
\hline
Modified-MobileFaceNet V2        & 2-5 MP               & 92.23              & \uline{2}      & 96.11              & \textbf{1}      & 95.15              & 4              & 95.88              & 4                   & 99.58              & 4             & 93.95                  & 6             & 32          & \textit{3}               & 456.89                                   & 7             & 8.4           & 5                        & 2.10                 & 4.33        & 5                        \\ 
\hline
ShuffleNetv2x2.0                 & 2-5 MP               & 89.27              & 8              & 92.71              & 8               & 90.88              & 9              & 88.08              & 9                   & 99.03              & 9             & 80.92                  & 8             & 3           & 9                        & 310.92                                   & 6             & 20.00         & 9                        & 4.97                 & 0.65        & 9                        \\ 
\hline\hline
MobileFaceNet Q8-bit \cite{DBLP:conf/icpr/BoutrosDK22}            & Baseline             & 87.95              & -              & 91.40              & -               & 95.05              & -              & 95.47              & -                   & 99.43              & -             & 90.57                  & -             & -           & -                        & 933                                      & -             & 1.10          & -                        & 1.1                  & -           & -                        \\ 
\hline
MobileFaceNet Q6-bit \cite{DBLP:conf/icpr/BoutrosDK22}            & Baseline             & 84.57              & -              & 87.69              & -               & 93.30              & -              & 93.03              & -                   & 98.87              & -             & 83.13                  & -             & -           & -                        & 933                                      & -             & 0.79          & -                        & 1.1                  & -           & -                        \\ 
\hline
PocketNetM-256 \cite{DBLP:journals/access/BoutrosSKDKK22}                  & Baseline             & 90.03              & -              & 95.66              & -               & 95.63              & -              & 97.17              & -                   & 99.58              & -             & 92.70                  & -             & -           & -                        & 1099.15                                  & -             & 7.0           & -                        & 1.75                 & -           & -                        \\ 
\hline
PocketNetM-128 \cite{DBLP:journals/access/BoutrosSKDKK22}                  & Baseline             & 90.00              & -              & 95.07              & -               & 95.67              & -              & 96.78              & -                   & 99.65              & -             & 92.63                  & -             & -           & -                        & 1099.02                                  & -             & 6.74          & -                        & 1.68                 & -           & -                        \\ 
\hline 
Idiap EdgeFace-XS($\gamma$=0.6) \cite{george2023edgeface}  & 2 MP                 & 91.88              & \textbf{1}     & 94.46              & \textit{3}      & 95.25              & \textbf{1}     & 95.72              & \textbf{\textbf{1}} & 99.68              & \textbf{1}    & 94.78                  & \textbf{1}    & 39          & \textbf{1}               & 153.99                                   & 5             & 7.17          & 7                        & 1.77                 & 5.0         & \textbf{1}               \\ 
\hline
Idiap EdgeFace-XXS-Q \cite{george2023edgeface}            & 2 MP                 & 89.65              & 5              & 93.11              & 5               & 94.68              & 4              & 93.77              & 4                   & 99.50              & 4             & 92.97                  & 4             & 22          & 4                        & 94.72                                    & \textit{3}    & 1.73          & \uline{2}                & 1.24                 & 3.92        & 4                        \\ 
\hline
MobileNet$_\text{V2-visteam}$    & 2 MP                 & 82.90              & 8              & 89.39              & 7               & 88.63              & 6              & 83.65              & 7                   & 98.58              & 6             & 51.60                  & 7             & 7           & 7                        & 86.20                                    & \uline{2}     & 3.38          & \textit{3}               & 1.70                 & 2.17        & 6                        \\ 
\hline
SAM-MFaceNet eHWS V1             & 2 MP                 & 91.35              & \uline{2}      & 95.01              & \textbf{1}      & 95.10              & \uline{2}      & 95.57              & \textit{3}          & 99.55              & \textit{3}    & 93.07                  & \uline{2}     & 35          & \uline{2}                & 236.75                                   & 6             & 4.4           & 4                        & 1.10                 & 4.68        & \uline{2}                \\ 
\hline
SAM-MFaceNet eHWS V2             & 2 MP                 & 91.28              & \textit{3}     & 94.73              & \uline{2}       & 94.90              & \textit{3}     & 95.72              & \textbf{1}          & 99.65              & \uline{2}     & 93.06                  & \textit{3}    & 33          & \textit{3}               & 236.75                                   & 6             & 4.4           & 4                        & 1.10                 & 4.45        & 3                        \\ 
\hline
SQ-HH                            & 2 MP                 & 84.13              & 6              & 91.60              & 6               & 87.17              & 7              & 84.28              & 6                   & 98.07              & 7             & 63.36                  & 6             & 10          & 6                        & 1399.39                                  & 8             & 4.55          & 6                        & 1.20                 & 1.47        & 8                        \\ 
\hline
ShuffleNetv2x0.5                 & 2 MP                 & 83.48              & 7              & 87.76              & 8               & 86.00              & 8              & 80.33              & 8                   & 97.72              & 8             & 38.57                  & 8             & 1           & 8                        & 17.14                                    & \textbf{1}    & 0.77          & \textbf{1}               & 0.17                 & 1.92        & 7                        \\ 
\hline
ShuffleNetv2x1.5                 & 2 MP                 & 89.73              & 4              & 93.44              & 4               & 91.08              & 5              & 88.78              & 5                   & 98.95              & 5             & 77.11                  & 5             & 20          & 5                        & 147.21                                   & 4             & 7.90          & 8                        & 1.99                 & 2.78        & 5                       
\end{tabular}
}
\vspace{1mm}
\caption{Evaluation results of the submitted solutions and baselines on the benchmarks introduced in Section \ref{sec:eval_datasets}. Also, FLOPs, model size, and number of parameters are given. For each dataset, the respective achieved rank for each submission is given. The Borda count and rank over all verification benchmarks are given in the Accuracy column. The respective ranking is given for FLOPs and model size. The combined, final ranking is a weighted Borda count of the achieved accuracy ($70\%$), the FLOPs ranking ($15\%$) and the model size ($15\%$). Submissions with 2-5M parameters are labeled as "2-5 MP". Submissions with $<2$M parameters are labeled as "2 MP".}
\label{tbl:overall_result_with_borda_count}
\vspace{8mm}
\end{table*}

\begin{table*}
\centering
\resizebox{\textwidth}{!}{%
\begin{tabular}{c|c|cccc|ccc||c||cc||cccc}
\multicolumn{1}{c}{}             & \multicolumn{1}{c}{} & \multicolumn{7}{c||}{\textbf{RFW}}                                                                                       & \multicolumn{3}{c||}{\textbf{Low-Resolution}}                                                                                                                                 & \multicolumn{4}{c}{\textbf{IJB-C (TAR @ FAR)}}                                                     \\
\textbf{Model}                   & \textbf{Category}    & \textbf{Asian} & \textbf{African} & \textbf{Caucasian} & \textbf{Indian} & \textbf{Avg.} & \textbf{Std.} & \textbf{SER} & \textbf{XQLFW} & \begin{tabular}[c]{@{}c@{}}\textbf{TinyFace}\\\textbf{(Rank 1)}\end{tabular} & \begin{tabular}[c]{@{}c@{}}\textbf{TinyFace}\\\textbf{(Rank 5)}\end{tabular} & \textbf{FAR=$10^{-5}$} & \textbf{FAR=$10^{-4}$} & \textbf{FAR=$10^{-3}$} & \textbf{FAR=$10^{-2}$}  \\ 
\hline
ResNet-18 ArcFace \cite{arcface}               & Baseline             & 93.30          & 94.03            & 97.33              & 95.25           & 94.98         & 1.53          & 2.51         & 78.82          & 55.95                                                                        & 61.96                                                                        & 90.39                  & 93.58                  & 95.92                  & 97.61                   \\ 
\hline
ResNet-50 ArcFace \cite{arcface}               & Baseline             & 97.07          & 98.15            & 99.07              & 98.08           & 98.09         & 0.71          & 3.14         & 81.32          & 62.23                                                                        & 66.57                                                                        & 93.49                  & 95.67                  & 97.22                  & 98.23                   \\ 
\hline
ResNet-100~ElasticFace-Arc \cite{DBLP:conf/cvpr/BoutrosDKK22}      & Baseline             & 98.77          & 99.27            & 99.50              & 98.92           & 99.11         & 0.29          & 2.47         & 82.30          & 64.51                                                                        & 69.17                                                                        & 94.43                  & 96.50                  & 97.62                  & 98.32                   \\ 
\hline
ResNet-100~ElasticFace-Cos  \cite{DBLP:conf/cvpr/BoutrosDKK22}     & Baseline             & 98.75          & 99.30            & 99.52              & 99.03           & 99.15         & 0.29          & 2.59         & 83.77          & 65.82                                                                        & 69.74                                                                        & 94.57                  & 96.51                  & 97.67                  & 98.42                   \\ 
\hline\hline
EfficientNet$_\text{b0-visteam}$ & 2-5 MP               & 86.03          & 85.28            & 91.20              & 87.52           & 87.51         & 2.28          & 1.67         & 88.05          & 52.70                                                                        & 58.79                                                                        & 73.64                  & 85.04                  & 92.73                  & 96.97                   \\ 
\hline
GhostFaceNetV1-1 KU \cite{ghostfacenets}             & 2-5 MP               & 95.10          & 96.03            & 98.30              & 96.40           & 96.46         & 1.16          & 2.88         & 84.72          & 59.54                                                                        & 64.75                                                                        & 92.22                  & 94.93                  & 96.76                  & 97.96                   \\ 
\hline
GhostFaceNetV1-2 KU \cite{ghostfacenets}             & 2-5 MP               & 93.88          & 94.80            & 97.15              & 95.35           & 95.30         & 1.19          & 2.15         & 83.73          & 55.87                                                                        & 61.13                                                                        & 90.95                  & 94.06                  & 96.25                  & 97.78                   \\ 
\hline
Idiap EdgeFace-S($\gamma$=0.5) \cite{george2023edgeface}   & 2-5 MP               & 94.75          & 95.85            & 97.67              & 95.18           & 95.86         & 1.11          & 2.25         & 89.95          & 61.02                                                                        & 65.47                                                                        & 92.42                  & 95.63                  & 97.42                  & 98.49                   \\ 
\hline
Idiap EdgeFace-XS-Q \cite{george2023edgeface}             & 2-5 MP               & 91.90          & 92.30            & 95.92              & 92.52           & 93.16         & 1.61          & 1.98         & 88.08          & 59.20                                                                        & 64.75                                                                        & 90.80                  & 94.40                  & 96.68                  & 98.02                   \\ 
\hline
MB2-HH                           & 2-5 MP               & 81.87          & 81.05            & 89.97              & 84.00           & 84.22         & 3.49          & 1.89         & 76.68          & 59.25                                                                        & 67.03                                                                        & 68.22                  & 79.86                  & 89.31                  & 95.52                   \\ 
\hline
Modified-MobileFaceNet V1        & 2-5 MP               & 92.30          & 92.98            & 96.68              & 93.20           & 93.79         & 1.70          & 2.32         & 87.75          & 64.29                                                                        & 69.60                                                                        & 90.01                  & 93.99                  & 96.60                  & 98.23                   \\ 
\hline
Modified-MobileFaceNet V2        & 2-5 MP               & 92.27          & 92.90            & 96.63              & 93.42           & 93.80         & 1.68          & 2.30         & 87.80          & 64.24                                                                        & 69.74                                                                        & 89.88                  & 93.95                  & 96.55                  & 98.22                   \\ 
\hline
ShuffleNetv2x2.0                 & 2-5 MP               & 80.97          & 80.48            & 89.35              & 83.03           & 83.46         & 3.53          & 1.83         & 78.32          & 38.62                                                                        & 46.72                                                                        & 69.25                  & 80.92                  & 89.96                  & 95.93                   \\ 
\hline\hline
Idiap EdgeFace-XS($\gamma$=0.6) \cite{george2023edgeface}  & 2 MP                 & 93.35          & 93.98            & 96.95              & 94.28           & 94.64         & 1.37          & 2.18         & 88.15          & 58.77                                                                        & 63.89                                                                        & 90.80                  & 94.78                  & 96.92                  & 98.24                   \\ 
\hline
Idiap EdgeFace-XXS-Q  \cite{george2023edgeface}           & 2 MP                 & 89.85          & 90.30            & 94.60              & 92.13           & 91.72         & 1.87          & 1.88         & 87.60          & 57.40                                                                        & 62.44                                                                        & 87.55                  & 92.97                  & 96.08                  & 97.90                   \\ 
\hline
MobileNet$_\text{V2-visteam}$    & 2 MP                 & 79.20          & 82.90            & 87.97              & 82.22           & 83.07         & 3.15          & 1.73         & 85.60          & 44.90                                                                        & 52.19                                                                        & 30.21                  & 51.60                  & 74.07                  & 89.93                   \\ 
\hline
SAM-MFaceNet eHWS V1             & 2 MP                 & 90.38          & 92.10            & 96.12              & 92.78           & 92.85         & 2.08          & 2.48         & 86.08          & 61.31                                                                        & 66.33                                                                        & 87.83                  & 93.07                  & 95.98                  & 97.86                   \\ 
\hline
SAM-MFaceNet eHWS V2             & 2 MP                 & 90.87          & 91.82            & 96.05              & 92.82           & 92.89         & 1.95          & 2.31         & 85.98          & 61.07                                                                        & 66.36                                                                        & 87.80                  & 93.06                  & 95.97                  & 97.82                   \\ 
\hline
SQ-HH                            & 2 MP                 & 74.43          & 71.63            & 84.50              & 78.62           & 77.30         & 4.85          & 1.83         & 69.10          & 52.09                                                                        & 61.13                                                                        & 47.84                  & 63.36                  & 78.26                  & 89.96                   \\ 
\hline
ShuffleNetv2x0.5                 & 2 MP                 & 73.93          & 72.28            & 83.60              & 76.92           & 76.68         & 4.32          & 1.69         & 69.95          & 33.39                                                                        & 41.79                                                                        & 4.68                   & 38.57                  & 73.82                  & 91.75                   \\ 
\hline
ShuffleNetv2x1.5                 & 2 MP                 & 81.08          & 80.95            & 89.50              & 82.33           & 83.47         & 3.52          & 1.81         & 76.53          & 31.73                                                                        & 39.88                                                                        & 36.09                  & 77.11                  & 89.75                  & 96.13                  
\end{tabular}
}

\caption{Evaluation results of the submitted solutions on the RFW, XQLFW, and IJB-C benchmarks. For RFW, the verification accuracy is given for Asian, African, Caucasian, and Indian subgroups. The average accuracy and the standard deviation are calculated from the accuracies of the four ethnicity datasets. The SER value indicates the bias of a model, with a higher value indicating a higher bias of the model. For XQLFW, the verification accuracy is given. For TinyFaces, rank 1 and rank 5 metrics are shown. For IJB-C the TAR at different FAR are shown. Additionally, baselines are given for all datasets. Submissions with 2-5M parameters are labeled as "2-5 MP". Submissions with $<2$M parameters are labeled as "2 MP".}
\label{tbl:evaluation}
\vspace{25mm}
\end{table*}

{
\bibliographystyle{ieee}

}

\end{document}